\pdfoutput=1

\documentclass[11pt]{article}

\usepackage[final]{acl}
\usepackage{xcolor}
\usepackage{graphicx} 
\usepackage{subcaption} 
\usepackage{booktabs}
\usepackage{float} 
\usepackage{stfloats} 
\usepackage{array} 
\usepackage{amsthm}
\usepackage{amssymb}
\usepackage{amsmath}
\usepackage{mathtools} 

\usepackage{geometry}
\usepackage{caption}

\usepackage{times}
\usepackage{latexsym}

\usepackage[T1]{fontenc}

\usepackage[utf8]{inputenc}

\usepackage{microtype}

\usepackage{inconsolata}

\usepackage{graphicx}

\title{Automated Tone Transcription and Clustering with Tone2Vec}

\author{
 \textbf{Yi Yang\textsuperscript{1}},
 \textbf{Yiming Wang\textsuperscript{1}},
 ZhiQiang Tang\textsuperscript{2},
 \textbf{Jiahong Yuan\textsuperscript{1}}
\\
 \textsuperscript{1}University of Science and Technology of China, \textsuperscript{2}Anhui University
\\
 \texttt{\{yanggnay, wangyiming\}@mail.ustc.edu.cn} \\ \texttt{774564456@qq.com}, \texttt{jiahongyuan@ustc.edu.cn}
}

\begin{document}
\maketitle

\begin{abstract}

Lexical tones play a crucial role in Sino-Tibetan languages. However, current phonetic fieldwork relies on manual effort, resulting in substantial time and financial costs. This is especially challenging for the numerous endangered languages that are rapidly disappearing, often compounded by limited funding. In this paper, we introduce pitch-based similarity representations for tone transcription, named \texttt{Tone2Vec}. Experiments on dialect clustering and variance show that \texttt{Tone2Vec} effectively captures fine-grained tone variation. Utilizing \texttt{Tone2Vec}, we develop the first automatic approach for tone transcription and clustering by presenting a novel representation transformation for transcriptions. Additionally, these algorithms are systematically integrated into an open-sourced and easy-to-use package, \href{https://github.com/YiYang-github/ToneLab}{\texttt{ToneLab}}, which facilitates automated fieldwork and cross-regional, cross-lexical analysis for tonal languages. Extensive experiments were conducted to demonstrate the effectiveness of our methods. 

\end{abstract}

\section{Introduction}

As the second-largest language family in the world, the Sino-Tibetan languages comprise over 400 languages, nurturing the cultural and communicative bonds of 1.4 billion speakers (Wikipedia). Given the prevalence of lexical tones in most Sino-Tibetan languages~\citep{thurgood2003sino}, phonetic fieldwork typically involves conducting tone transcription for each word in the survey lexicon across unexplored regions, followed by categorizing these transcriptions into the respective tone categories of the region. Exploring lexical tones enriches both linguistic and historical research, including migration patterns~\citep{mig1}, contact between languages~\citep{cont1}, and their evolution over time~\citep{evo1, evo2, evo3}.

However, existing methodologies face two primary obstacles that hinder further investigation, research, and documentation of Sino-Tibetan languages. 

\vspace{-2mm}

\begin{enumerate}
    \item \textbf{Obstacles in Documenting.} In practice, tone transcription relies on manual effort, and the recorders involved must undergo extensive and prolonged training, which typically lasts several months. Subsequently, the tone categories of a region are discerned based on these transcriptions. The absence of an automatic tone transcription and clustering system leads to substantial time and financial costs, especially for the vast number of endangered languages that are rapidly disappearing~\citep{hale1992endangered}, often with limited funding.
    \item \textbf{Obstacles in Analysis.} Although tones can be transcribed using a five-scale system, analyzing tones across different regions is challenging due to the varying lengths (2 or 3 units) of these transcriptions and the differing number of tones in each area. Moreover, extensive fieldwork, represented by the \href{http://www.china-language.gov.cn/}{Chinese Language Resources Protection Project}, has gathered abundant tone transcription data—exceeding one million records—from thousands of dialect regions within the Sino-Tibetan language family. This has created an urgent need to develop comparable features for different tone transcriptions and to use computational methods to analyze variations across these dialect regions.    

\end{enumerate}

\begin{figure*}[t]
\centering
\includegraphics[width=0.9 \textwidth]{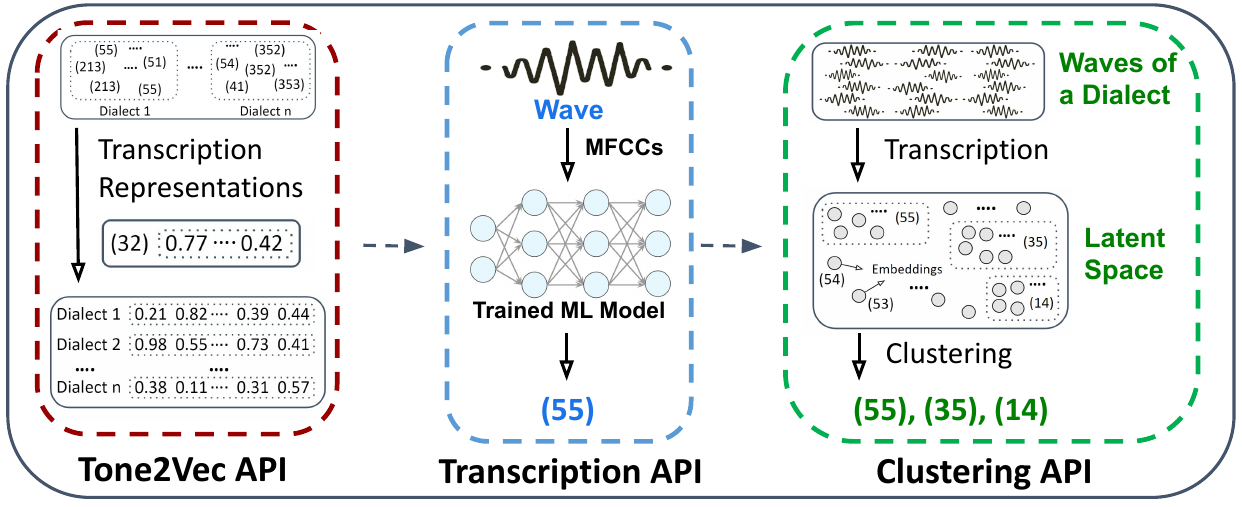}
\caption{Overview of our proposed methods. From left to right: \texttt{Tone2Vec} module for representations, Transcription module for automated tone transcription, and Clustering module for clustering tonal data. }
\vspace{-2mm}
\label{fig:tones_overview}
\end{figure*}

In this paper, we systematically addressed the above problems from three angles: feature construction, algorithm design, and the development of an easy-to-use tool. As illustrated in Figure~\ref{fig:tones_overview}, our contributions can be summarized as follows:

\begin{itemize}
    \item \textit{Our first contribution} is the proposal of  \texttt{Tone2Vec}, which maps diverse tone transcriptions to a comparable feature space. \texttt{Tone2Vec} constructs pitch-based similarity representations by mapping each transcription to a simulated smooth pitch variation curve. We also propose methods to construct tonal representations for dialect regions. By analyzing these representations across different dialect areas, we show that \texttt{Tone2Vec} captures tonal variations and clusters dialects more accurately than methods that treat each tone as an isolated category. 
    \item \textit{As our second contribution}, we developed the first automated algorithms for tone transcription and clustering. These algorithms are especially beneficial for endangered tonal languages. Experiments demonstrate that our models perform well in cross-regional tone transcription with less than 1,500 samples. Notably, our algorithms can accurately cluster tones using fewer than 60 speech samples for a given dialect. 
    \item \textit{As our third contribution}, all these algorithms are systematically integrated into \href{https://github.com/YiYang-github/ToneLab}{\texttt{ToneLab}},  a user-friendly platform designed for both lightweight fieldwork and subsequent analysis in Sino-Tibetan Tonal Languages. Users can choose to use pretrained models or train new models with their own data for different scenarios. Researchers can also leverage \texttt{ToneLab} to propose new computational methods and conduct evaluations.
\end{itemize}

\section{Related Work}

\subsection{Representations}

Traditionally, linguists represent tones as discrete sequences using transcription systems like the Five-Scale, Four-Domain, Nine-Scale, and Contour Tone Marking Systems. Recent machine learning successes involve distilling complex entities like words, graphs, and speeches into computable, comparable representations, typically as multi-dimensional vectors, like word2vec~\citep{word2vec} and graph2vec~\citep{graph2vec}. In contrast to treating different tones as atomic units, \texttt{Tone2Vec} offers fine-grained tonal representations for tone transcriptions and tone analysis.

\subsection{Automated Tone Classification}

In recent years, automated tone classification methods~\citep{Ryant2014Mandarin,chen2016tone,yuan2021automatic,Alex2020Wav2Vec,yuanimproved} have achieved accuracy rates surpassing those of human listeners, nearing 100\% in Standard Mandarin. One approach involves preprocessing the raw signals into features using mel frequency cepstral coefficients (MFCCs), followed by classification prediction using models such as SVM~\citep{Ryant2014Mandarin}, MLP~\citep{Ryant2014Mandarin}, and Convolutional Neural Networks (CNNs)~\citep{chen2016tone}. Another strategy~\citep{yuan2021automatic, yuanimproved} leverages more powerful pre-trained models like Wav2Vec 2.0~\citep{Alex2020Wav2Vec} for fine-tuning. However, tone classification methods require prior categorical information, are primarily applied on Standard Mandarin, and cannot transcribe tones across dialects.

\section{Preliminary}

\subsection{Lexical Tones}

\begin{figure}[h]
\centering
\includegraphics[width= \columnwidth]{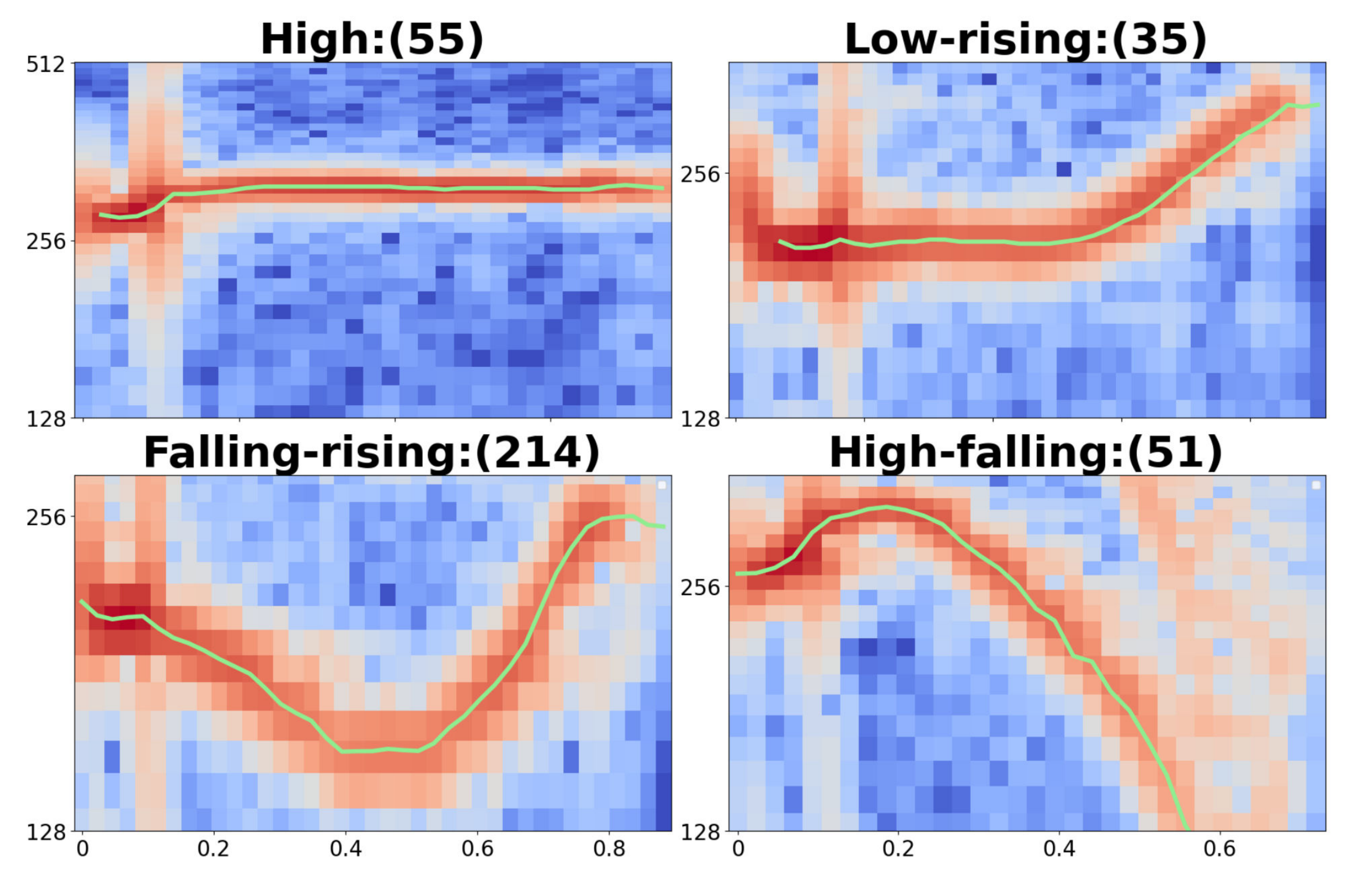}
\caption{Fundamental frequency (F0, represented with solid lines) and transcription (e.g., \texttt{(55)} indicating a High tone) for the four basic Mandarin tones.}
\label{fig:tones}
\end{figure}

In tonal languages such as Standard Mandarin, lexical meanings are differentiated by pitch variations. These lexical tones are annotated using a scale from 1 (lowest) to 5 (highest), in accordance with Chao's Tone Letter system~\citep{chao1930system}. The four basic lexical tones and pitch variations are visually expounded in Figure~\ref{fig:tones} by the fundamental frequency, F0.

\subsection{Five-scale Marking System}\label{subsec:five-degree}

The Five-scale Marking System, developed by Yuen-Ren Chao \citep{chao1930system}, is the most widely used method for transcribing tones in the Sino-Tibetan language family. In this system, the pitch of a person's speech is divided into five relative levels: \texttt{(1)}, \texttt{(2)}, \texttt{(3)}, \texttt{(4)}, and \texttt{(5)}, where \texttt{(1)} indicates the lowest pitch and \texttt{(5)} the highest. Tones are then transcribed using sequences of two or three numbers to represent the pitch contour over time. For example, a tone that starts at the mid-level pitch and rises to the high level might be transcribed as \texttt{(35)}. The relative changes between these numbers indicate the pitch movement. For example, the tones \texttt{(53)} and \texttt{(42)} both represent a falling pitch, but the first starts at the highest level \texttt{(5)} and ends at a mid-level \texttt{(3)}, while the second starts one level lower, beginning at \texttt{(4)} and ending at \texttt{(2)}. It is worth noting that transcription represents relative pitch, not absolute pitch. Different speakers may produce the same relative pitch at different absolute levels; for example, one person's lowest pitch might not be the same as another's, but listeners can still identify it as the lowest pitch in their speech~\citep{honorof2005}.

\subsection{Tone Classification, Transcription and Clustering Tasks}
\label{subsec:task_def}

Let \( S(t) \) be a speech signal and \( T = \ <n_1, n_2, \ldots, n_k> \) as the corresponding transcription, where \( t \) represents time. We denote a set of speech signals from a dialectal region as \( \mathcal{S} = [ S_{1}(t), S_{2}(t), \ldots, S_{m}(t) ] \), where each \( S_{i}(t) \) represents a speech signal,

\textbf{Tone Classification Task}: Given a dialect area with a certain number of tone categories, for instance, there are M categories, the tone classification task \(l \) can be defined as shown in Equation~\ref{eq:classi_def}.

\begin{equation}\label{eq:classi_def}
    \begin{aligned}
        l: \mathcal{S} &= [ S_{1}(t), S_{2}(t), \ldots, S_{m}(t) ] \\
        \rightarrow \mathcal{T} &= [ t_{1}, t_{2}, \ldots, t_{m} ], \\
        t_{i} &= \{1, 2, \ldots, m\} \\
    \end{aligned}
\end{equation}

\textbf{Tone Transcription Task}: Unlike tone classification, the tone Transcription task \( f \)  takes speech from any dialect as input and outputs a five-scale transcription rather than categories. This process can be defined as shown in Equation~\ref{eq:trans_def}.

\begin{equation}\label{eq:trans_def}
    \begin{aligned}
    f: S(t) &\rightarrow T = \ <n_1, n_2, \ldots, n_k>, \\
    n_i &\in \{1, 2, 3, 4, 5\}, \\
    k &\in \{2, 3\}
    \end{aligned}
\end{equation}

Note that, without any prior knowledge (e.g., speaker's highest/lowest pitch, all tone categories), it is hard to distinguish between a level tone \texttt{(55)} and a level tone \texttt{(44)}, or a \texttt{(41)} and a \texttt{(51)} from a single speech signal. However, tones like \texttt{(523)} and \texttt{(51)} can be distinguished due to their different variations. In our subsequent tone evaluation, we will also take this into account, using only the relative pitch as the criterion for assessment. 

\textbf{Tone Clustering Task}: The objective of the tone clustering task \( g \) is to group these signals into N distinct tonal categories \(\mathcal{T} = [ T_{1}, T_{2}, \ldots, T_{N} ] \), defined as Equation~\ref{eq:cluster_def}, where N is not known and needs model automatic judgment.

\begin{equation}\label{eq:cluster_def}
    \begin{aligned}
        g: \mathcal{S} &= [ S_{1}(t), S_{2}(t), \ldots, S_{m}(t) ] \\
        \rightarrow \mathcal{T} &= [ T_{1}, T_{2}, \ldots, T_{N} ], \\
        T_{i} &= \ <n_{i, 1}, n_{i, 2}, \ldots, n_{i, k(i)}\ >, \\
    n_{i,j} &\in \{1, 2, 3, 4, 5\}, \\
    k(i) &\in \{2, 3\}
    \end{aligned}
\end{equation}

\section{Data}

The majority of publicly available speech data labeled for tones are limited to the four tone categories (T1-T4) in standard Mandarin~\citep{RyuTonePerfect,bu2017aishell}.  There is a lack of comprehensive, cross-regional speech data transcribed using the five-scale marking system. To address these limitations, we managed to collect a speech dataset to develop models for automatic tone transcription and clustering, and a second, transcription-only dataset to demonstrate the application of the \texttt{ToneLab} tone analysis tool. 

Both datasets are in Jianghuai Mandarin. which boasts approximately 70 million speakers and has been extensively studied~\cite{tang-dissertation,zeng2018case}. Jianghuai Mandarin contains many dialect regions that differ from each other in their tonal systems ~\cite{complexity1, complexity2, complexity3}. With its rich tonal resources, Jianghuai Mandarin serves as a valuable testbed for training and evaluating tone transcription and clustering systems, especially at an early stage where open-source speech with five-scale tone transcription labels is scarce.

Below, we provide a detailed introduction and preprocessing steps for the two datasets.

\textbf{2238 Recordings from 11 Jianghuai Mandarin Dialects (\texttt{Dataset1})}: We managed to compile a carefully curated dataset from a previous study\cite{tang-dissertation}, which includes 2238 speech recordings across 11 Jianghuai Mandarin dialects. Each speech sample was transcribed by experienced Sino-Tibetan linguists using the five-scale marking system. The dataset categorizes speakers into four groups for each dialect: young males(\texttt{YM}), young females(\texttt{YF}), older males(\texttt{OM}), and older females(\texttt{OF}). Tone clusterings are meticulously defined for each group in every region. Each Jianghuai Mandarin dialect is accompanied by detailed descriptions of geographical locations, tone classifications, and dialect regions, all detailed in Appendix~\ref{section:jianghuai_info}. In subsequent experiments, we randomly selected data from 7 regions for training, 2 regions for validation, and 2 regions for testing, out of a total of 11 regions. The best-performing parameters on the validation set were then used for the final test set evaluation.

\textbf{Transcriptions with Dialect Cluster Labels (\texttt{Dataset2})}: In the study of Chinese tones, \texttt{Hongchao} and \texttt{Huangxiao} clusters of dialect regions in Jianghuai Mandarin are often used to investigate tone evolution, such as the lengthening of entering tones~\citep{tang-dissertation}, tone sandhi~\citep{complexity2, coblin2005comparative}, and tonal inventories~\citep{complexity2}. We obtained transcription data from 19 dialect areas in the \texttt{Hongchao} cluster and 12 dialect areas in the \texttt{Huangxiao} cluster from the \href{http://www.china-language.gov.cn/}{Chinese Language Resources Protection Project}, which is the largest language resource database in the world. Each dialect area includes 1000 tone transcriptions from the same survey word list, totaling 31,000 transcriptions. Detailed information is provided in Appendix~\ref{section:jianghuai_info}.

\section{Tone2Vec: From Tones to Vectors}
\label{sec:method}

 In this section, we propose pitch-based similarity representations by quantifying the differences in pitch variations inherent in tones, which we call \texttt{Tone2Vec}. \texttt{Tone2Vec} is an easy-to-use, simple, and effective method for measuring similarity distance. \texttt{Tone2Vec} not only enables the comparison of tonal variations across dialects but also provides a straightforward loss function for training automatic tone transcription and clustering models.

\subsection{From Categories to Pitch-based Similarity Representations}

\begin{figure}[t]
\centering
\includegraphics[width=\columnwidth]{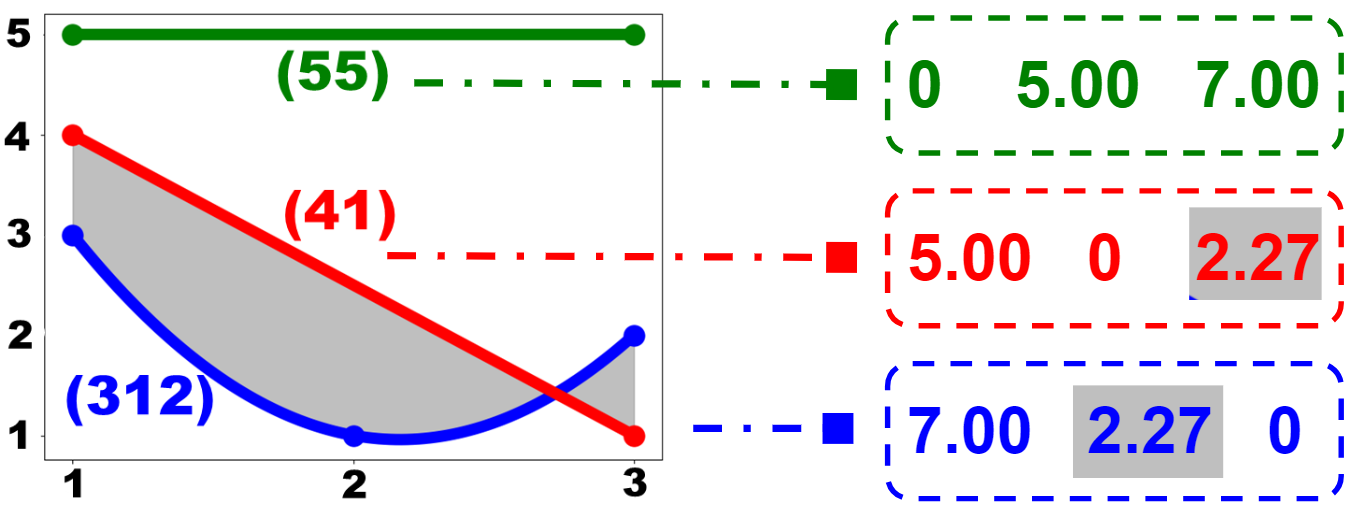}
\vspace{-5.5mm}
\caption{\textbf{Left}: Visual simulations using transcription sequences \( l_{1} \) = \texttt{(55)} (green linear curve), \( l_{2} \) = \texttt{(41)} (red linear curve), and \( l_{3} \) = \texttt{(312)} (blue quadratic curve). Grey shading denotes the area between \texttt{(41)} and \texttt{(312)}. \textbf{Right}: The number 2.27 with grey shading represents the calculated distance between \texttt{(41)} and \texttt{(312)}.}
\vspace{-2mm}
\label{fig:trans2vec}
\end{figure}

In \texttt{Tone2Vec}, we map each transcription $l$, such as \texttt{(55)}, to a simulated smooth pitch variation curve $p_{l}(x)$. As shown in Figure~\ref{fig:trans2vec}, for transcriptions with two units, a linear curve is employed to represent pitch variations, while for those of three units, such as \texttt{(312)}, we employ a quadratic curve to smoothly interpolate the points \((1, 3)\), \((2, 1)\), and \((3, 2)\). The divergence between any pair of tone transcriptions, \(l_1\) and \(l_2\), is quantitatively assessed by calculating the area between their pitch variation curves, expressed as \(D(l_1, l_2) = \int_{[1,3]} |f_{l_1}(x) - f_{l_2}(x)| dx\). This measure quantifies the differences in pitch variations. 
Given $n$ transcription sequences $l_{1},...,l_{n}$, we can construct a $n \times n$ distance matrix $\mathcal{C} = {(D(l_{i}, l_{j}))}_{i,j} \in \mathbb{R}^{n \times n}$, where each row represents the features of a transcription, capturing the subtle pitch variation differences among them.

\subsection{Case Study: Dialect Clustering and Variance}

To better introduce and prove the effectiveness of our methods, we conducted experiments on Dialect Group Clustering and Variance using \texttt{Dataset2}. The Dialect Clustering task involves classifying 31 dialect regions, each with 1,000 transcription entries, into two clusters, and the metric accuracy is reported. The task of dialect variance aims to quantify the differences between dialect regions. A good representation should hierarchically reflect dialect variance. We compared \texttt{Tone2Vec} with the baseline model, \texttt{Baseline}. For \texttt{Baseline}, the difference between two transcriptions is 0 if they are identical, and 1 otherwise.

For the dialect clustering task, we calculated the average transcription difference for each pair of dialect areas to derive their tonal features, then performed clustering and evaluated the accuracy of the predicted labels against the true labels. To account for the influence of clustering techniques, we employed seven different methods following the study~\citep{bartelds2022quantifying}: single link (\texttt{sl}), complete link (\texttt{cl}), group average (\texttt{ga}), weighted average (\texttt{wa}), unweighted centroid (\texttt{uc}), weighted centroid (\texttt{wc}), and minimum variance (\texttt{mv}) clustering~\citep{cluster1,clustering2}. The best results are reported in Table~\ref{tab:dialect_accuracy} and the results of all seven methods are available in Appendix~\ref{section:appendix_case}.

\begin{figure}[H]
    \centering
    \begin{minipage}[b]{0.9\columnwidth}
        \centering
        \includegraphics[width=\textwidth]{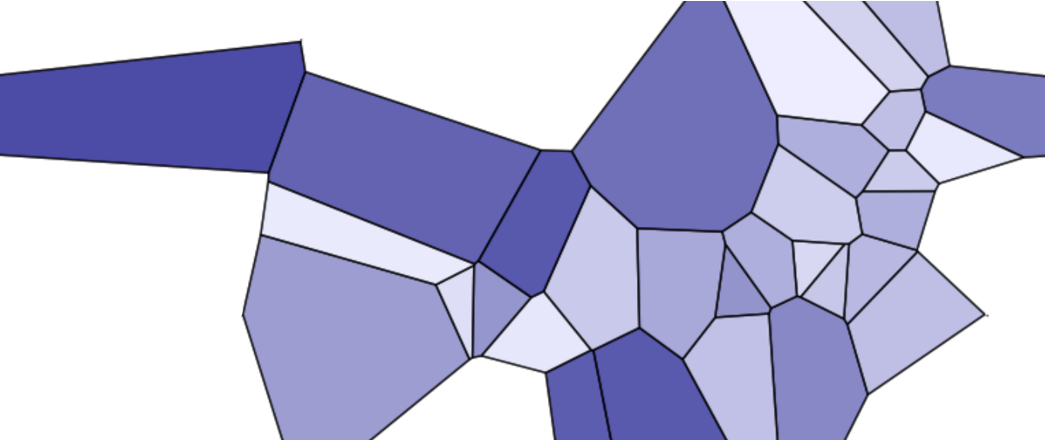} 
        \vspace{-8mm}
        \caption*{(a) \texttt{Tone2Vec} }
    \end{minipage}
    \begin{minipage}[b]{0.9\columnwidth}
        \centering
        \includegraphics[width=\textwidth]{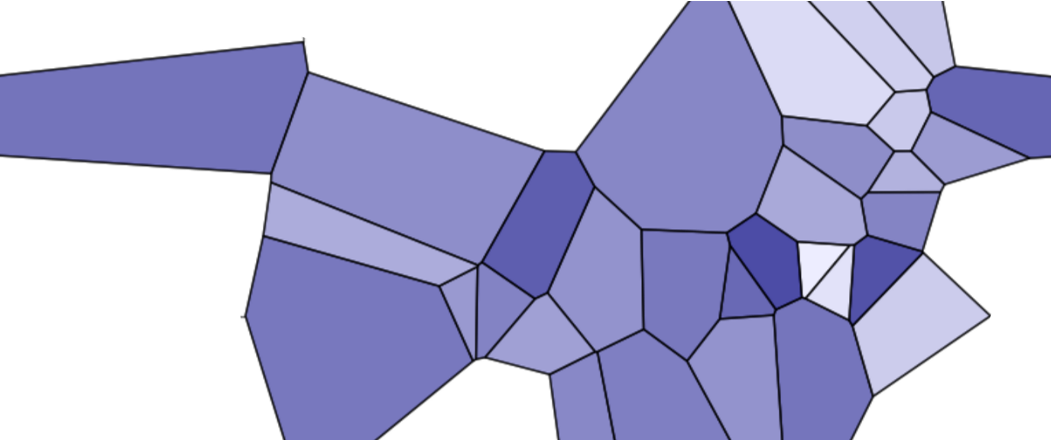} 
        \vspace{-8mm}
        \caption*{(b) \texttt{Category} }
    \end{minipage}
    \vspace{-3mm}
    \caption{MDS maps visualizing pronunciation differences across dialects. Similar colors indicate similar pronunciations.}
    \label{fig:dialect_accuracy_mds}
\end{figure}

For the dialect variance task, we use multidimensional scaling (MDS)~\citep{torgerson1952multidimensional, bartelds2022quantifying} to reduce the dimensionality of the dialect representations to 1. The value differences between regions intuitively reflect the variance across different areas and are depicted with varying color intensities in Figure~\ref{fig:dialect_accuracy_mds}.

\begin{figure}[t]
    \centering
    \begin{minipage}[b]{0.9\columnwidth}
        \centering
        \includegraphics[width=\textwidth]{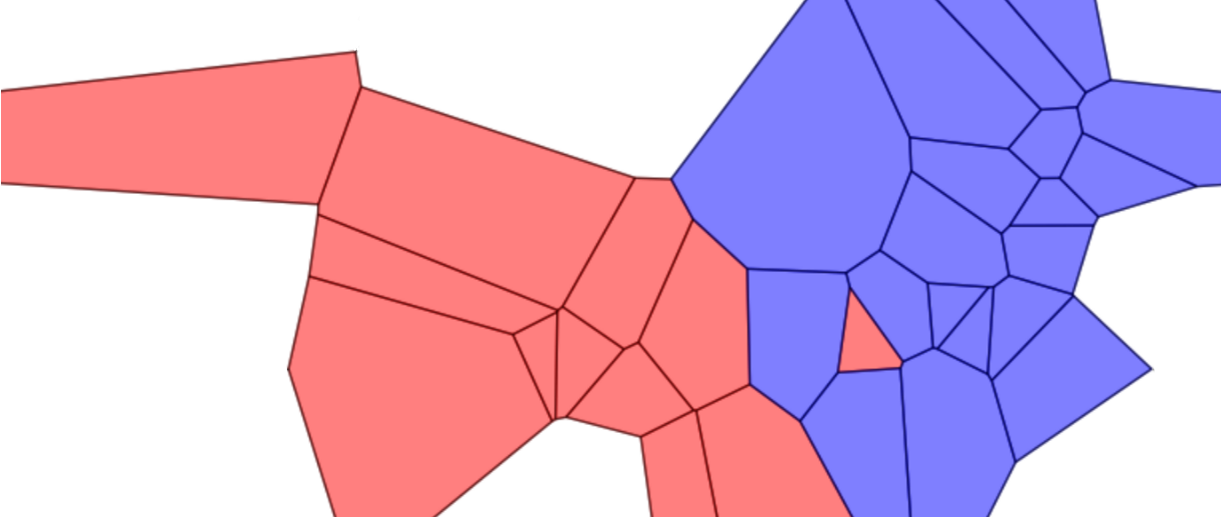} 
        \vspace{-8mm}
        \caption*{(a) Gold-standard}
    \end{minipage}
    \begin{minipage}[b]{0.9\columnwidth}
        \centering
        \includegraphics[width=\textwidth]{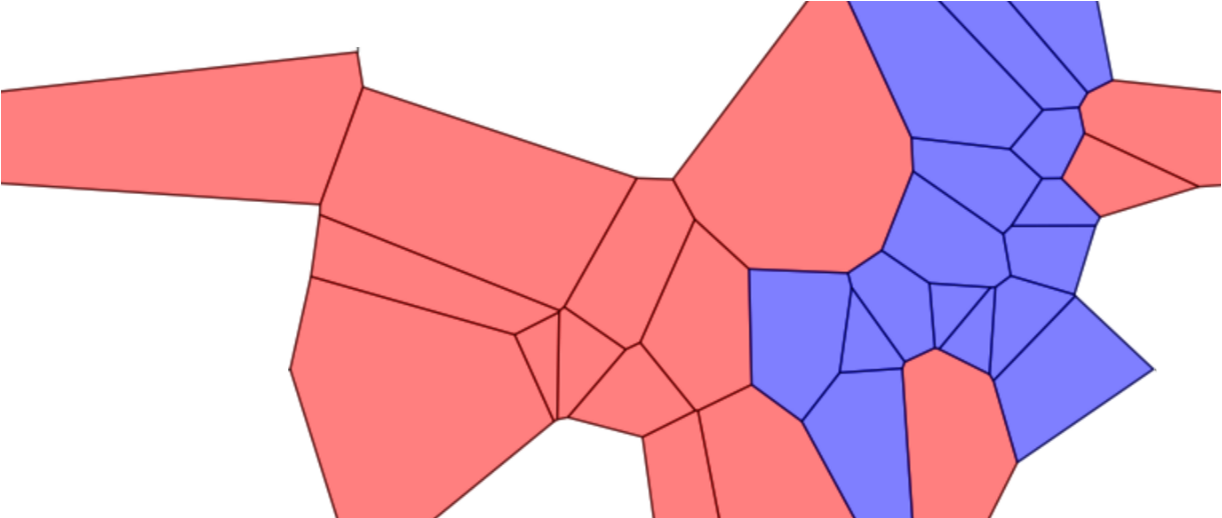} 
        \vspace{-8mm}
        \caption*{(b) \texttt{Tone2Vec} with \texttt{mv} clustering}
    \end{minipage}
    \begin{minipage}[b]{0.9\columnwidth}
        \centering
        \includegraphics[width=\textwidth]{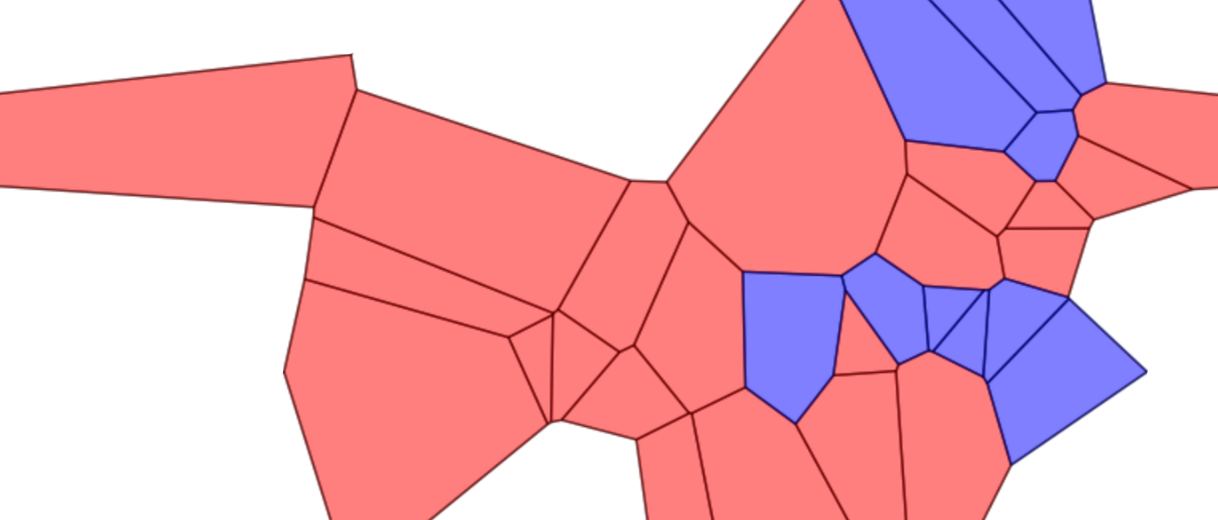} 
        \vspace{-8mm}
        \caption*{(c) \texttt{Category} with \texttt{wa} clustering}
    \end{minipage}
    \vspace{-3mm}
    \caption{Cluster maps visualizing the \texttt{Huangxiao} and \texttt{Hongchao} dialect clusters. Red represents \texttt{Huangxiao} and blue represents \texttt{Hongchao}.}
    \label{fig:dialect_accuracy_visual}
\end{figure}

\textbf{Discussion} The accuracy results in Table~\ref{tab:dialect_accuracy} show that \texttt{Tone2Vec} outperforms the \texttt{Baseline} by 12.90\%. Additionally, the visualization in Figure~\ref{fig:dialect_accuracy_visual} indicates that clustering constructed by \texttt{Tone2Vec} is more balanced, whereas the \texttt{Baseline} method tends to classify most dialect areas into a single cluster. Figure~\ref{fig:dialect_accuracy_mds} demonstrates that \texttt{Tone2Vec} better captures dialect variation, while the \texttt{baseline} method is more influenced by outliers, resulting in most areas having colors within a smaller range.

\begin{table}[h]
\centering 
\vspace{-2mm}
\resizebox{0.45\textwidth}{!}{
    \begin{tabular}{l c c }
    \toprule
    \textbf{Method} & \textbf{Accuracy (\%)} & \textbf{Clustering} \\
    \midrule
    \texttt{Baseline} & 70.97 & \texttt{wa} \\
    \texttt{Tone2Vec} & \underline{83.87} & \texttt{mv} \\
    \bottomrule
    \end{tabular}
}
\caption{Accuracy of \texttt{Tone2Vec} and \texttt{Baseline} method in Dialect Group Clustering with the best clustering method. The underlined value represents the higher accuracy.}
\vspace{-4mm}
\label{tab:dialect_accuracy}
\end{table}

\section{Automatic Tone Transcription}

\subsection{Pitch-based Loss Function}

In contrast to CTC's explicit handling of transcriptions with variable lengths~\citep{Alex2006CTC}, our model \(\mathcal{M}\) implicitly discerns the length of the transcription sequence during the inference stage. We first fix the model \(\mathcal{M}\)'s output to consistently produce three float points. For each training instance \(x_{j}\), the model yields an output \(z_{j} = (z_{j,1}, z_{j,2}, z_{j,3})\), where every \(z_{i}\) falls within the pitch range [1,5]. When viewed through the lens of pitch variations, a sequence of length two—whether a level tone like \texttt{(55)}, an ascending tone like \texttt{(35)}, or a descending tone like \texttt{(53)}—exhibits a linear relationship among the three predicted components \(\mathcal{M}(x_{0}) = z_{0} = (z_{0,1}, z_{0,2}, z_{0,3})\). Sequences of length three, characteristic of contour tones such as \texttt{(352)} or \texttt{(334)}, lack this linearity. By establishing a threshold \(\beta\), we can determine the linearity of a sequence. For speech data \(x_{0}\), the inferred transcription \(\hat{y_{0}}\) can be formulated as shown in Equation~\ref{eq:tone_prediction}:

\begin{equation}
\label{eq:tone_prediction}
\hat{y_{0}} =
\begin{cases}
\begin{aligned}
(\lfloor z_{0,1} \rceil, \lfloor z_{0,3} \rceil) & \\
\text{if } | z_{0,1} + z_{0,3} - 2 \times z_{0,2} | &< \beta, \\
(\lfloor z_{0,1} \rceil, \lfloor z_{0,2} \rceil, \lfloor z_{0,3} \rceil) & \quad \text{otherwise}.
\end{aligned}
\end{cases}
\end{equation}

Here, \(\lfloor \rceil\) denotes the operation of rounding to the nearest whole number. The default value for $\beta$ is 0.5.

Building on \texttt{Tone2Vec}, we propose a pitch-based loss function, designated \(\mathcal{L}_{pitch}\), to automate the transcription of tones and represent signals as tonal representations. By recognizing that each numeral in a transcription sequence, ranging from 1 to 5, symbolizes a different pitch level, and the metric \(D(l_1, l_2)\) mirrors the discrepancy between sequences, the metric itself can be directly employed as the loss function for training. For simplicity, we use the mean absolute error (MAE) loss \(\hat{D}(\mathcal{M}(x_{j}), y_{j})\), which approximates \(D(\mathcal{M}(x_{j}), y_{j})\) in Equation~\ref{eq:example}.

\begin{equation}
\label{eq:example}
    \mathcal{L}_{pitch}(\mathcal{X},\mathcal{Y}) = -\sum_{j=1}^{N} \hat{D}(\mathcal{M}(x_{j}), y_{j})
\end{equation}

To introduce this concept more intuitively, We denote \(\mathcal{M}(x_{j})\) as \((z_{j,1}, z_{j,2}, z_{j,3})\). If \(y_{j}\) is a sequence of length three, i.e., \((y_{j,1}, y_{j,2}, y_{j,3})\), then the distance \(\hat{D}(\mathcal{M}(x_{j}), y_{j})\) is defined as:

\begin{equation}
\begin{aligned}
\hat{D}(\mathcal{M}(x_{j}), y_{j}) = &|z_{j,1} - y_{j,1}| + |z_{j,2} - y_{j,2}|  \\
&+ |z_{j,3} - y_{j,3}| \\
\end{aligned}
\end{equation}

If \(y_{j}\) is a sequence of length three, i.e., \((y_{j,1}, y_{j,2}, y_{j,3})\), then the distance \(\hat{D}(\mathcal{M}(x_{j}), y_{j})\) is defined as:

\begin{equation}
\begin{aligned}
\hat{D}(\mathcal{M}(x_{j}), y_{j}) = &|z_{j,1} - y_{j,1}| + |z_{j,3} - y_{j,2}| \\
&+ |z_{j,2} - \frac{1}{2} (y_{j,1} + y_{j,2})| 
\end{aligned}
\end{equation}

The selection of metric \(D\) centers on capturing the nuances of pitch variations inherent in tones. In this paper, we map each transcription $l$ to a simulated smooth pitch variation curve $f_{l}(x)$.

\subsection{Experiments}
\label{subsec:transcription_exp}

The experiments were conducted using \texttt{Dataset1}. In the absence of a baseline, we noted that linguists could record tone transcriptions by observing the fundamental frequency (F0) curves (Figure~\ref{fig:tones}), as indicated by ~\cite{chen2016tone}. 

We use quadratic fitting to regress twenty evenly sampled points from the F0 curve, using the values regressed from the second, middle, and second-to-last points as the predicted tone sequence. We first normalize these values and then use Equation~\ref{eq:tone_prediction} to infer the transcription. Although this method is not a standard automatic tone transcription system (since none currently exists), using F0 curves is a common practice in tone research.

Since the absolute pitch of a speaker is difficult to derive from single-syllable speech alone, we propose a new metric, \texttt{Variance}, to describe the average discrepancy between model predictions and labeled transcriptions through normalized pitch variation. First, we normalize any transcription \( l \) within the range [0, 1], denoted as \( f_{1}(l) \). Specifically, we map the highest pitch value to 1, and the lowest to 0, and evenly distribute the intermediate values. The examples below illustrate our process:

\begin{itemize}
    \item Transcription \texttt{(412)}:
    \begin{align*}
        \text{max: } 4, \text{ min: } 1 &\rightarrow \left(\frac{\texttt{4} - 1}{4 - 1}, \frac{\texttt{1} - 1}{4 - 1}, \frac{\texttt{2} - 1}{4 - 1}\right) \\
        &= (1, 0, 0.333)
    \end{align*}
    \item Transcription \texttt{(25)}:
    \begin{align*}
        \text{max: } 5, \text{ min: } 2 &\rightarrow \left(\frac{\texttt{2} - 1}{5 - 2}, \frac{\texttt{5} - 1}{5 - 2}\right) = (0, 1)
    \end{align*}
\end{itemize}

For any two transcriptions, \( l_{1} \) and \( l_{2} \), we obtain their relative pitches \( f_{1}(l_{1}) \) and \( f_{1}(l_{2}) \). We use \(\hat{D}(\sigma(f_{1}(l_{1})), \sigma(f_{1}(l_{2})))\) to measure the difference in relative pitch, resulting in the \texttt{Variance} metric, where \(\sigma\) is the sigmoid function. Lower variance indicates better model performance. For a more intuitive presentation, Table~\ref{tab:example_variance} shows the \texttt{Variance} values for the transcription \texttt{(445)} compared to six other transcriptions.

\begin{table}[h]
\centering
\vspace{-2mm}
\resizebox{0.47\textwidth}{!}{
    \begin{tabular}{l c | c c | c c}
    \toprule
    \textbf{Seq.} & \texttt{Variance} & \textbf{Seq} &  \texttt{Variance} & \textbf{Seq} &  \texttt{Variance} \\
    \midrule
    \texttt{(445)} & 0.0000 & \texttt{(45)} & 0.1225 & \texttt{(245)} & 0.1608 \\
     \texttt{(255)} & 0.2311 &  \texttt{(154)} & 0.2829 & \texttt{(251)} & 0.5243  \\
    \bottomrule
    \end{tabular}
}
\caption{\texttt{Variance} values for transcription \texttt{(445)} compared to \texttt{(45)}, \texttt{(245)}, \texttt{(255)}, \texttt{(154)} and \texttt{(251)}.}
\vspace{-4mm}
\label{tab:example_variance}
\end{table}

We tested our method on three models: ResNet~\cite{he2015resnet}, VGG~\cite{VGG}, and DenseNet~\cite{huang2017densely}. The model selection is informed by many previous studies~\cite{gao2019tonenet,chen2016tone} indicating that CNN models perform well in tone classification. Hyperparameters, such as the learning rate, were selected through grid search. Signals were preprocessed using Mel Frequency Cepstral Coefficients (MFCCs) before training the models. Each result is based on three separate experiments, and the averages are reported.

\begin{table}[h]
\centering
\vspace{-2mm}
\resizebox{0.45\textwidth}{!}{
    \begin{tabular}{l c c c}
    \toprule
    \textbf{Model} & \textbf{Method} & \textbf{Accuracy (\%)} & \texttt{Variance} \\
    \midrule
      & F0 & 10.07 & 0.2165 \\
    \midrule
    ResNet & \texttt{Tone2Vec} & 55.99 & 0.1222 \\
    VGG & \texttt{Tone2Vec} & \underline{56.08} & \textbf{0.1052} \\
    DenseNet & \texttt{Tone2Vec} & \textbf{61.01} & \underline{0.1083} \\
    \bottomrule
    \end{tabular}
}
\caption{Accuracy and \texttt{variance} of tone transcription using F0 extraction and \texttt{Tone2Vec} on ResNet, VGG, and DenseNet models. Higher accuracy or lower variance indicates better model performance. The bold value represents the best result, and the underlined value represents the second-best result.}
\vspace{-4mm}
\label{tab:transcription}
\end{table}

\textbf{Discussion} As illustrated in Table~\ref{tab:transcription}, our automatic tone transcription method significantly outperforms the F0 extraction-based approach in both Accuracy and \texttt{Variance} metrics. Combined with the examples in Table~\ref{tab:example_variance}, our model maintains consistently high performance across three models, with DenseNet showing the best in Accuracy and the VGG model excelling in \texttt{Variance}. These findings collectively indicate that using \texttt{Tone2Vec} to train models for automatic tone transcription effectively captures pitch variations.

\section{Automatic Tone Clustering}

\subsection{Clustering on Transcription Features}

Many studies~\citep{yuanimproved, pepino2021emotion, zerveas2021transformer} have shown that well-trained machine learning models not only perform well on targeted tasks but also provide hierarchical embeddings. Therefore, by extracting intermediate layer features, the automatic tone transcription model \(\mathcal{M}\), has already assigned tonal representations for each speech instance. Hence, the task of Tone Clustering can be regarded as a clustering task on transcription features. We then employ the clustering algorithm DBSCAN~\citep{ester1996density} on these representations to determine the number of tone categories automatically, selecting the most probable predicted label in each cluster as a tone category.

\subsection{Experiments}

\begin{table}[h]
\centering
\resizebox{0.45\textwidth}{!}{
    \begin{tabular}{lcccc c}
    \toprule
    \textbf{SPK} & \textbf{Type} & \textbf{Tone 1} & \textbf{Tone 2} & \textbf{Tone 3} & \textbf{Tone 4} \\
    \midrule
    \texttt{OF} & Lab.  & \texttt{(213)} & \texttt{(24)} & \texttt{(41)} & \texttt{(53)} \\
     & Pred. & \texttt{(313)} & \texttt{(45)} & \texttt{(51)} & \texttt{(42)} \\
     \midrule
    \texttt{YF} & Lab.  & \texttt{(212)} & \texttt{(24}) & \texttt{(51)} & \texttt{(55)} \\
     & Pred. & \texttt{(213)} & \texttt{(34)} & \texttt{(52)} & \texttt{(44)} \\
     \midrule
    \texttt{OM} & Lab.  & \texttt{(213)} & \texttt{(24)} & \texttt{(41)} & \texttt{(51)} \\
     & Pred. & \texttt{(212)} & \texttt{(34)} & \texttt{(31)} & \texttt{(32)} \\
    \bottomrule
    \end{tabular}
}
\caption{Comparison of manually labelled (Lab.) and automatically predicted (Pred.) tone categories for young females (\texttt{YF}), older males (\texttt{OM}), and older females (\texttt{OF}) in the \texttt{Wuhu} dialect area. Pred values indicate the transcriptions, with each non-dash value representing a predicted category.}
\label{tab:tone_clustering_wuhu}
\end{table}

\begin{table}[h]
\centering
\resizebox{0.45\textwidth}{!}{
    \begin{tabular}{lcccc c}
    \toprule
    \textbf{SPK} & \textbf{Type} & \textbf{Tone 1} & \textbf{Tone 2} & \textbf{Tone 3} & \textbf{Tone 4} \\
    \midrule
    \texttt{YM} & Lab.  & \texttt{(41)} & \texttt{(24)} & \texttt{(31)} & \texttt{(55)} \\
     & Pred. & \texttt{-} & \texttt{(24)} & \texttt{(32)} & \texttt{(44)} \\
     \midrule
    \texttt{OF} & \texttt{Lab.}  & \texttt{(41)} & \texttt{(24)} & \texttt{(31)} & \texttt{(55)} \\
     & Pred. & \texttt{(41)} & \texttt{-} & \texttt{(212)} & \texttt{(45)} \\
     \midrule
    \texttt{YF} & Lab.  & \texttt{(51)} & \texttt{(24)} & \texttt{(32)} & \texttt{(55)} \\
     & Pred. & \texttt{(51)} & \texttt{(24)} & \texttt{(43)} & \texttt{(33)} \\
     \midrule
    \texttt{OM} & Lab.  & \texttt{(41)} & \texttt{(24)} & \texttt{(32)} & \texttt{(55)} \\
     & Pred. & \texttt{(52)} & \texttt{(23)} & \texttt{(31)} & \texttt{(44)} \\
    \bottomrule
    \end{tabular}
}
\caption{Comparison of manually labelled (Lab.) and automatically predicted (Pred.) tone categories for young males (\texttt{YM}), young females (\texttt{YF}), older males (\texttt{OM}), and older females (\texttt{OF}) in the \texttt{Yangzhou} dialect area. Pred values indicate the transcriptions, with each non-dash value representing a predicted category.}
\label{tab:tone_clustering_yangzhou}
\end{table}

The experiments were conducted using \texttt{Dataset1}. We still use the 7:2:2 data split strategy for training and model selection, following the transcription experiments in Subsection~\ref{subsec:transcription_exp}. Each region has at most four clusterings from four speakers: young males (\texttt{YM}), young females (\texttt{YF}), older males (\texttt{OM}), and older females (\texttt{OF}). Each speaker's speech, consisting of fewer than 60 samples per dialect, is manually labeled for tone categories. We select the best-performing model, \texttt{DenseNet}, for tone transcription tasks. Tonal embeddings are visualized using UMap~\citep{mcinnes2020umap}, with DBSCAN parameters \texttt{eps} set to 0.6 and \texttt{min\_samples} set to 4.

\begin{figure}[H]
    \centering
    \begin{minipage}[b]{0.9\columnwidth}
        \centering
        \includegraphics[width=\textwidth]{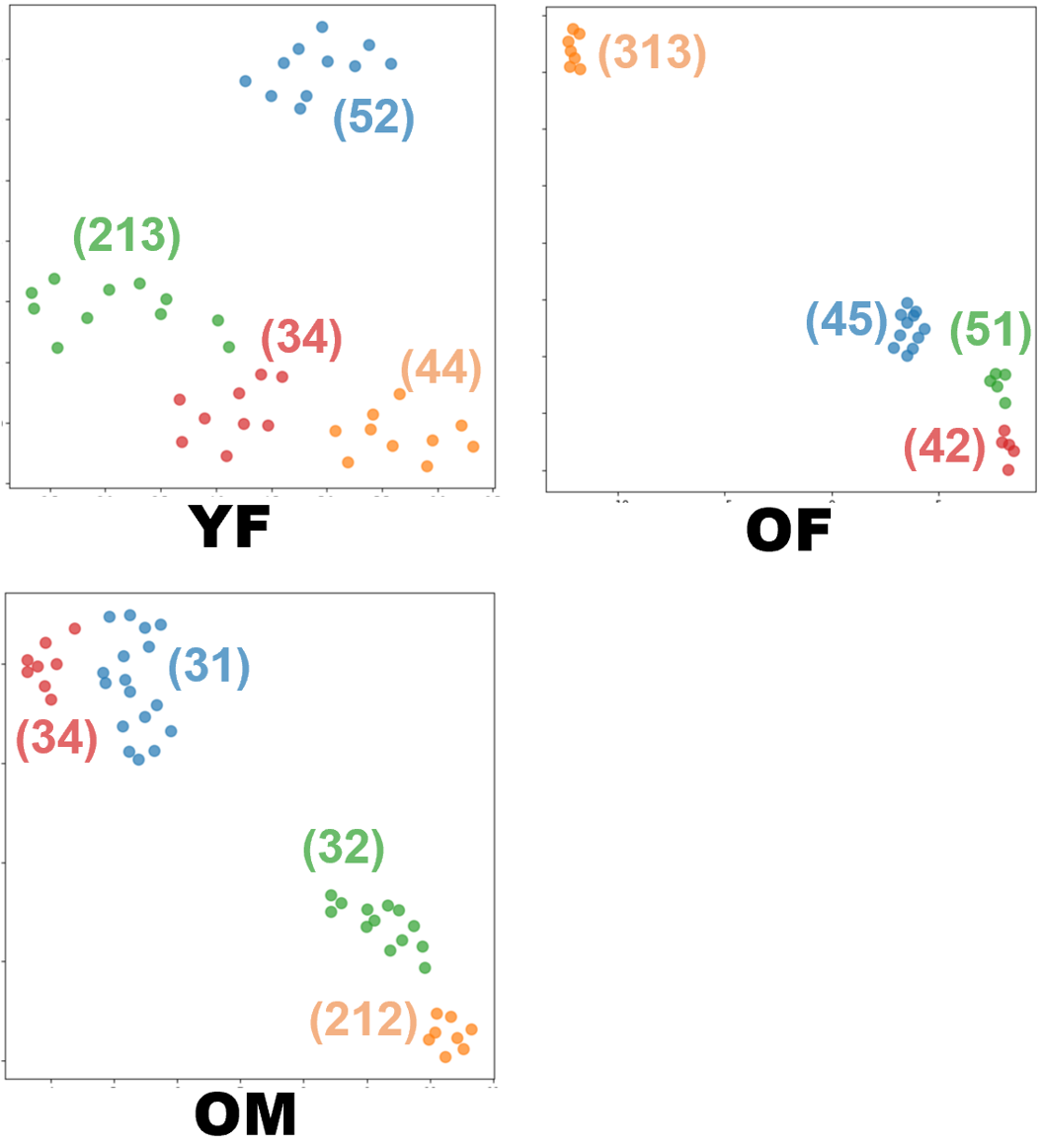} 
        \vspace{-8mm}
        \caption*{(a) \texttt{Wuhu} }
    \end{minipage}
    \begin{minipage}[b]{0.9\columnwidth}
        \centering
        \includegraphics[width=\textwidth]{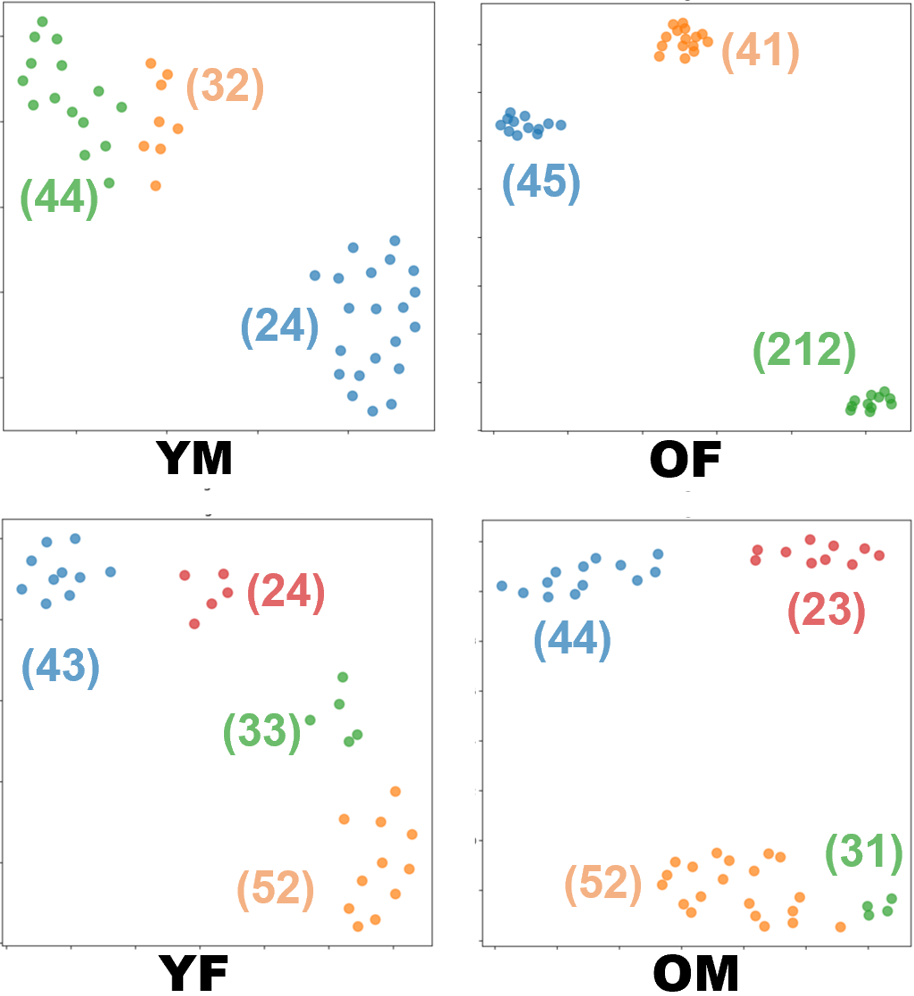} 
        \vspace{-8mm}
        \caption*{(b) \texttt{Yangzhou} }
    \end{minipage}
    \vspace{-3mm}
    \caption{Visualization of automatic clustering for young females (\texttt{YF}), older males (\texttt{OM}), and older females (\texttt{OF}) in the \texttt{wuhu} dialect areas and young females (\texttt{YF}), older males (\texttt{OM}), and older females (\texttt{OF}) in the \texttt{wuhu} dialect using UMAP for dimensionality reduction and DBSCAN for clustering.}
    \label{fig:cluster_visual}
\end{figure}

\textbf{Discussion} As illustrated in Table~\ref{tab:tone_clustering_wuhu} and Table~\ref{tab:tone_clustering_yangzhou}, our model accurately determined the number of tone categories with 71\% accuracy. Additionally, the model generally predicted rising tones as rising, falling tones as falling, and contour tones as contour. Differences between predictions and ground truth mainly stemmed from variations in pitch magnitude, such as predicting \texttt{(212)} as \texttt{(213)}. Overall, these differences are within an acceptable margin. Notably, tone categorization varies among different individuals. Simultaneously, as depicted in Figure ~\ref{fig:cluster_visual}, tonal features show clear clustering. The proximity of \texttt{(52)} to \texttt{(31)} rather than to \texttt{(23)} reflects inner similarities among different tones.

\section{ToneLab: A User-friendly Platform for Tonal Languages}

We have developed an easy-to-use package, \href{https://github.com/YiYang-github/ToneLab}{\texttt{ToneLab}}. We aim for \texttt{ToneLab} to be a user-friendly platform for lightweight documentation and quantitative analysis in Sino-Tibetan tonal languages. To sum up, two main modules are introduced.

\subsection{Automatic Tone Documentation Solutions}

This module supports automatic tone transcription and clustering  for studying tonal languages. 


\textbf{Input}: MFCCs extracted from speech, either one (for transcription and lightweight classification) or multiple (for clustering).

\textbf{Models}: MLP and CNN models, including ResNet~\cite{he2015resnet}, VGG~\cite{VGG}, and DenseNet~\cite{huang2017densely}. Users can use the provided models or train their own models with their own data.

\subsection{Quantative Cross Dialect Tone Analysis}

In \texttt{ToneLab}, representations can be easily queried from the pre-computed database for any tone transcriptions. It supports inputting a set of transcriptions from a dialect region and returns the comparable tonal features of that region, which can be used to study dialect clustering and variance. Our package also supports investigating the influence of initials and finals on tones using methods such as the improved Levenshtein distance~\citep{wieling2012inducing}.

\section{Conclusion}

In this paper, we proposed Automated Tone Transcription and Clustering with \texttt{Tone2Vec}. We hope our work could raise awareness about the importance and urgency of preserving and studying endangered Sino-Tibetan tonal languages, which have long been overlooked, and encourages more collaborative efforts in this crucial field.

\section{Limitations}
\label{sec: limitation}

As an early exploratory work, this paper focuses solely on single syllables with CNN models. Further studies involving continuous speech recognition on powerful transformer-based speech models like \textit{wav2vec2.0}~\citep{Alex2020Wav2Vec} could be conducted. And, our models are currently built using only a few thousand labeled speech data points due to the limited open-sourced data. Additionally, we found that embeddings from the intermediate layers of trained transcription models effectively reflect tonal representations, though further considerations are needed to enhance phonological theories and phonetic analysis.

\section{Acknowledgments}
This work was partially supported by USTC Grant YD2110002303 and the Rose-New Lotus Scholars Program at the School of the Gifted Young.

\bibliography{acl_latex.bbl}

\newpage
\appendix

\section{Detailed Dialect Information}
\label{section:jianghuai_info}

Table~\ref{tab:appendix_t1} provides detailed information on the province, city, cluster, sub-slices, East Longitude, and North Latitude for the 31 dialect regions. The positions of the dialect regions in Figure~\ref{fig:dialect_accuracy_visual} and Figure~\ref{fig:dialect_accuracy_mds} are determined by their actual East Longitude and North Latitude.

\begin{table*}[h]
\centering
\begin{tabular}{ccccccc}
\toprule
Point & Province & City & Cluster & Sub-slices & East Longitude (°E) & North Latitude (°N) \\
\hline
1 & \texttt{Jiangxi} & \texttt{Jiujiang} & \texttt{Huangxiao} & \texttt{-} & 115.408 & 29.617 \\
2 & \texttt{Jiangxi} & \texttt{Jiujiang} & \texttt{Huangxiao} & \texttt{-} & 116.012 & {29.735} \\
3 & \texttt{Anhui} & \texttt{Tongling} & \texttt{Huangxiao} & \texttt{-} & {117.442} & {30.883} \\
4 & \texttt{Anhui} & \texttt{Anqing} & \texttt{Huangxiao} & \texttt{-} & {117.020} & {30.300} \\
5 & \texttt{Shaanxi} & \texttt{Shangluo} & \texttt{Huangxiao} & \texttt{-} & {109.160} & {33.429} \\
6 & \texttt{Hubei} & \texttt{Huanggang} & \texttt{Huangxiao} & \texttt{Luotian} & {115.433} & {30.925} \\
\texttt{7} & \texttt{Hubei} & \texttt{Xiaogan} & \texttt{Huangxiao} & \texttt{Xiaogan} & {113.533} & {30.925} \\
\texttt{8} & \texttt{Hubei} & \texttt{Xiaogan} & \texttt{Huangxiao} & \texttt{Yunmeng} & {113.759} & {31.027} \\
\texttt{9} & \texttt{Hubei} & \texttt{Xiaogan} & \texttt{Huangxiao} & \texttt{Xiaogan} & {113.817} & {31.733} \\
\texttt{10} & \texttt{Hubei} & \texttt{Huanggang} & \texttt{Huangxiao} & \texttt{E'dong} & {114.581} & {31.303} \\
\texttt{11} & \texttt{Hubei} & \texttt{Huanggang} & \texttt{Huangxiao} & \texttt{-} & {115.917} & {30.008} \\
12 & \texttt{Hubei} & \texttt{Xiaogan} & \texttt{Huangxiao} & \texttt{-} & {113.633} & {31.275} \\
 \midrule
13 & \texttt{Anhui} & \texttt{Chuzhou} & \texttt{Hongchao} & \texttt{Yangzhou} & {118.933} & {32.700} \\
14 & \texttt{Anhui} & \texttt{Chuzhou} & \texttt{Hongchao} & \texttt{-} & {118.312} & {32.301} \\
15 & \texttt{Anhui} & \texttt{Wuhu} & \texttt{Hongchao} & \texttt{-} & {118.408} & {31.258} \\
16 & \texttt{Anhui} & \texttt{Chizhou} & \texttt{Hongchao} & \texttt{Rongjiu} & {118.208} & \texttt{30.575} \\
17 & \texttt{Anhui} & \texttt{Xuancheng} & \texttt{Hongchao} & \texttt{-} & {119.350} & {30.908} \\
18 & \texttt{Anhui} & \texttt{Wuwei} & \texttt{Hongchao} & \texttt{-} & {117.908} & {31.217} \\
19 & \texttt{Anhui} & \texttt{Chizhou} & \texttt{Hongchao} & \texttt{-} & {117.467} & {30.525} \\
20 & \texttt{Anhui} & \texttt{Anqing} & \texttt{Hongchao} & \texttt{Anqing} & {116.908} & {30.958} \\
21 & \texttt{Anhui} & \texttt{Huainan} & \texttt{Hongchao} & \texttt{-} & {116.975} & {32.608} \\
22 & \texttt{Anhui} & \texttt{Xuancheng} & \texttt{Hongchao} & \texttt{-} & {119.117} & {31.133} \\
23 & \texttt{Anhui} & \texttt{Wuhu} & \texttt{Hongchao} & \texttt{-} & {118.508} & {31.175} \\
24 & \texttt{Anhui} & \texttt{Lu'an} & \texttt{Hongchao} & \texttt{Hongchao} & {116.633} & {31.675} \\
25 & \texttt{Jiangsu} & \texttt{Yancheng} & \texttt{Hongchao} & \texttt{-} & {120.205} & {33.396} \\
26 & \texttt{Jiangsu} & \texttt{Zhenjiang} & \texttt{Hongchao} & \texttt{-} & {119.430} & {32.195} \\
27 & \texttt{Jiangsu} & \texttt{Nanjing} & \texttt{Hongchao} & \texttt{-} & {118.460} & {32.020} \\
28 & \texttt{Jiangsu} & \texttt{Yangzhou} & \texttt{Hongchao} & \texttt{-} & {119.421} & {33.231} \\
29 & \texttt{Jiangsu} & \texttt{Yangzhou} & \texttt{Hongchao} & \texttt{-} & {119.430} & {32.380} \\
30 & \texttt{Jiangsu} & \texttt{Huai'an} & \texttt{Hongchao} & \texttt{-} & {119.375} & {33.883} \\
31 & \texttt{Jiangsu} & \texttt{Huai'an} & \texttt{Hongchao} & \texttt{-} & {119.032} & {33.559} \\
\bottomrule
\end{tabular}
\caption{Detailed Dialect Information from Hongchao and Huangxiao Clusters.}
\label{tab:appendix_t1}
\end{table*}

\section{Full Results of the Dialect Group Clustering}
\label{section:appendix_case}

Table~\ref{tab:dialect_accuracy_full} presents the results of seven clustering algorithms—single link (\texttt{sl}), complete link (\texttt{cl}), group average (\texttt{ga}), weighted average (\texttt{wa}), unweighted centroid (\texttt{uc}), weighted centroid (\texttt{wc}), and minimum variance (\texttt{mv})—applied to the \texttt{Tone2Vec} and \texttt{Baseline} methods.

\begin{table*}[h]
\centering
\resizebox{\textwidth}{!}{
    \begin{tabular}{l c c c c c c c}
    \toprule
    \textbf{Method} & \texttt{sl} & \texttt{cl} & \texttt{ga} & \texttt{wa} & \texttt{uc} & \texttt{wc} & \texttt{mv}  \\
    \midrule
    \texttt{Tone2Vec} & \underline{64.52} & \underline{70.97} & \underline{70.97} & 64.52 & \underline{70.97} & 61.29 & \underline{\textbf{83.87}} \\
    \texttt{Baseline} & 58.06 & 67.74 & 67.74 & \underline{\textbf{70.97}} & 61.29 & \underline{67.74} & 61.29 \\
    \bottomrule
    \end{tabular}
}
\caption{Accuracy of \texttt{Tone2Vec} and \texttt{Baseline} methods with all seven clustering algorithms in Dialect Group Clustering. The underlined values represent the higher accuracy for each clustering algorithm. Bold numbers represent the best performance for each method.}
\label{tab:dialect_accuracy_full}
\end{table*}

\textbf{Discussion} Table~\ref{tab:dialect_accuracy_full} indicates that the choice of clustering algorithm significantly affects accuracy, with a difference of 22.58\% between the best and worst clustering algorithms for \texttt{Tone2Vec} and 12.91\% for \texttt{Baseline}. Among the seven clustering algorithms, \texttt{Tone2Vec} outperformed \texttt{Baseline} in five methods, while \texttt{Baseline} outperformed in two. Considering the influence of different clustering algorithms, these results demonstrate that \texttt{Tone2Vec} provides better tone representations than \texttt{Baseline}, especially with the highest accuracy of 83.87\%, which is significantly higher than the best performance of \texttt{Baseline} at 70.97\%.

\end{document}